\documentclass[runningheads]{llncs}
\usepackage{amsmath}
\usepackage{amsfonts}
\usepackage{dsfont}
\usepackage{graphicx}
\usepackage{tikz}
\usepackage{multirow}
\usepackage{array}
\usepackage{tabularx,booktabs}
\usepackage{subcaption}
\usepackage{dirtytalk}
\usepackage{bm}
\usepackage{dsfont}
\usepackage{float}
\usetikzlibrary{arrows.meta}
\usepackage{color}
\usepackage{hyperref}
\usepackage{siunitx}
\usepackage{bbding}

\newcommand{\red}[1]{\textcolor{red}{#1}}
\newcommand{\green}[1]{\textcolor{green}{#1}}
\newcommand{\blue}[1]{\textcolor{blue}{#1}}
\newcommand{\gray}[1]{\textcolor{gray}{#1}}

\newcommand{\repeatthanks}{\textsuperscript{\thefootnote}}

\begin{document}

\title{Color Flow Imaging Microscopy Improves Identification of Stress Sources of Protein Aggregates in Biopharmaceuticals}

\titlerunning{Color Flow Imaging Microscopy Improves Stress-type Detection of SvPs}

\author{Michaela Cohrs\inst{1}\thanks{Equal contribution.} \and
Shiwoo Koak\inst{2}\repeatthanks \and
Yejin Lee\inst{2}\repeatthanks \and
Yu Jin Sung\inst{2}\repeatthanks \and\\
Wesley De Neve\inst{2, 3}  \and
Hristo L. Svilenov\inst{1} \and
Utku Ozbulak\inst{2, 3}
}

\authorrunning{Cohrs et al.}

\institute{
Faculty of Pharmaceutical Sciences, Ghent University, Ghent, Belgium
\and
Center for Biosystems and Biotech Data Science,\\ Ghent University Global Campus, Republic of Korea
\and
Department of Electronics and Information Systems,\\ Ghent University, Belgium\\
\email{utku.ozbulak@ghent.ac.kr}
}

\maketitle

\begin{abstract}
Protein-based therapeutics play a pivotal role in modern medicine targeting various diseases. Despite their therapeutic importance, these products can aggregate and form subvisible particles (SvPs), which can compromise their efficacy and trigger immunological responses, emphasizing the critical need for robust monitoring techniques. Flow Imaging Microscopy (FIM) has been a significant advancement in detecting SvPs, evolving from monochrome to more recently incorporating color imaging. Complementing SvP images obtained via FIM, deep learning techniques have recently been employed successfully for stress source identification of monochrome SvPs. In this study, we explore the potential of color FIM to enhance the characterization of stress sources in SvPs. To achieve this, we curate a new dataset comprising 16,000 SvPs from eight commercial monoclonal antibodies subjected to heat and mechanical stress. Using both supervised and self-supervised convolutional neural networks, as well as vision transformers in large-scale experiments, we demonstrate that deep learning with color FIM images consistently outperforms monochrome images, thus highlighting the potential of color FIM in stress source classification compared to its monochrome counterparts.
\end{abstract}

\section{Introduction}

Protein-based therapeutics have become critical in modern medicine for combating various human diseases. Currently, over 400 individual biopharmaceutical products hold active licenses from the U.S. Food and Drug Administration (FDA) and the European Medicines Agency (EMA), contributing to a notable market valued at greater than \$300 billion~\cite{mullard2021fda}. With the number of drugs approved for clinical use increasing dramatically over the past decade, the impact of protein-based treatments in sustaining modern healthcare is becoming more evident~\cite{walsh2022biopharmaceutical}. However, large biomolecules have delicate structures, making them prone to unfolding and aggregation~\cite{ripple2012protein}. This process is often promoted by stressors such as heat or mechanical stress encountered during different stages of the product life cycle~\cite{arakawa2000protection,kiese2008shaken,wang2005protein}. As such, protein-based therapeutics face challenges involving the presence of undesired particles. The emerging particles exhibit different sizes, ranging from the nanometer to the visible size range ($>$100 \textmu m), and can cause undesired efficacy loss and immune responses~\cite{moussa2016immunogenicity,thorlaksen2023vitro,zolls2012particles}. Therefore, it is crucial to implement thorough monitoring and control strategies to ensure the safety and efficacy of these therapeutics throughout their development, manufacturing, and storage processes~\cite{narhi2015subvisible}.

Quality assessment of protein-based therapeutics has reached a pivotal point with breakthroughs in imaging techniques. In particular, the development of Flow Imaging Microscopy (FIM) has enabled the highly sensitive detection of subvisible particles (SvP - particles of size 2 to 100 \textmu m), facilitating the assessment of particle count and size distribution, even for certain translucent protein particles~\cite{narhi2015subvisible,fim_nn}. Notably, FIM can detect and image more translucent protein particles than established methods such as light obscuration~\cite{matter2019variance,sharma2010micro}. The FIM images have recently been used in conjunction with deep learning (DL) algorithms to classify SvPs or to identify stress sources that cause SvPs~\cite{lai2022machine}. In this area, most research efforts employ grayscale images obtained with FIM due to prior limitations of the FIM technology. Nonetheless, recent advancements in FIM have enabled the capture of SvPs in color images through multiple optical lenses~\cite{thite2022machine}. However, the equipment to capture such images in color is more expensive and the benefit of color imaging for SvP classification is unclear.

In this study, we aim at addressing the following research question: is the color information in FIM images beneficial for deep learning methods in identifying stress sources of SvPs? To answer this research question, we curate a new dataset from commercial monoclonal antibody products that were subjected to heat and mechanical stress under controlled conditions to capture protein aggregate images using FIM. Utilizing state-of-the-art deep learning models and large-scale experiments, we perform stress-type classification and find that models trained with color FIM consistently outperform those trained with monochrome images. To the best of our knowledge, our study is the first in its category in evaluating the utility of color versus monochrome FIM images using deep learning methods. Our findings highlight the potential of color FIM in advancing the characterization and classification of stress sources in subvisible protein aggregates, contributing substantially to the field of biopharmaceutical quality control.

\begin{table}[t]
\centering
\scriptsize
\caption{For heat and mechanical stress, the number of images in the training and validation sets is given for each antibody and its corresponding treatment. Underlined mAbs are stressed with both heat and mechanical stress.}
\label{tbl:dataset}
\begin{tabular}{cccc}
\toprule
Stress type & Protein & \phantom{--}Training\phantom{--} & \phantom{--}Validation\phantom{--} \\
\midrule
\multirow{5}{*}{Heat stress} & \underline{mAb1} & 1,055 & 188 \\
~ & \underline{mAb2} &  2,298 & 409 \\
~ & mAb3 & 1,715 &  285 \\
~ & mAb4 & 1,332 & 227 \\
~ & mAb5 & 0 & 447\\
\midrule
\multirow{5}{*}{Mechanical stress} & \underline{mAb1} & 2,146 & 434 \\
~ & \underline{mAb2} & 1,071 & 200\\
~ & mAb6 & 1,580 & 311\\
~ & mAb7 & 1,603 & 288\\
~ & mAb8 & 0 & 367 \\
\bottomrule
\end{tabular}
\end{table}

\begin{figure}[t!]
    \centering
    \includegraphics[width=0.45\textwidth]{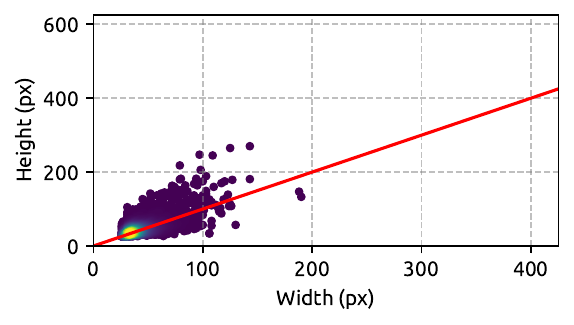}
    \includegraphics[width=0.45\textwidth]{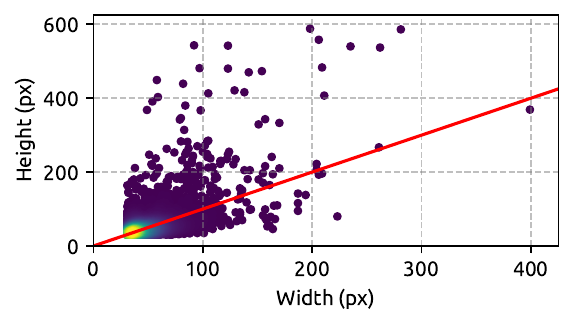}
    \caption{Distribution of the height and width of the SvP images used in this study for (left) heat stress and (right) mechanical stress. The red line in both figures highlights 1:1 aspect ratio.}
    \label{fig:scatter_plots}
\end{figure}

\section{Methodology}

In this section, we provide a description of the materials and the dataset used, outline the models utilized, and describe the color conversion and the training approach adopted.

\begin{figure}[t!]
\centering
\begin{subfigure}{0.98\textwidth}
\centering
\includegraphics[width=0.48\textwidth]{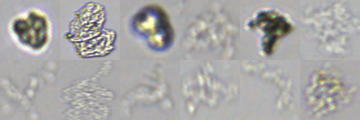}
\includegraphics[width=0.48\textwidth]{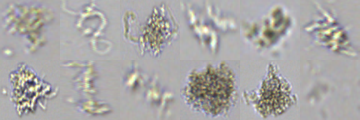}
\caption{RGB}
\end{subfigure}
\begin{subfigure}{0.98\textwidth}
\centering
\includegraphics[width=0.48\textwidth]{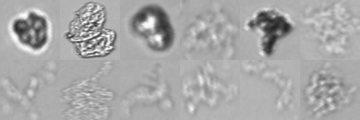}
\includegraphics[width=0.48\textwidth]{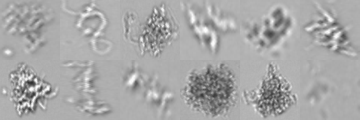}
\caption{Grayscale}
\end{subfigure}
\begin{subfigure}{0.98\textwidth}
\centering
\includegraphics[width=0.48\textwidth]{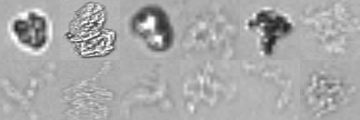}
\includegraphics[width=0.48\textwidth]{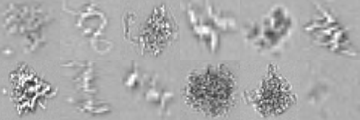}
\caption{Red channel}
\end{subfigure}
\begin{subfigure}{0.98\textwidth}
\centering
\includegraphics[width=0.48\textwidth]{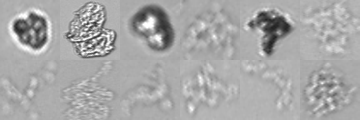}
\includegraphics[width=0.48\textwidth]{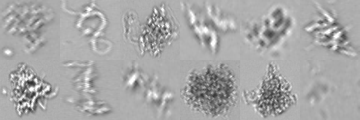}
\caption{Green channel}
\end{subfigure}
\begin{subfigure}{0.98\textwidth}
\centering
\includegraphics[width=0.48\textwidth]{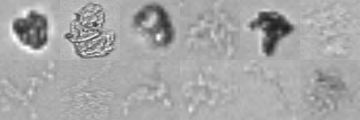}
\includegraphics[width=0.48\textwidth]{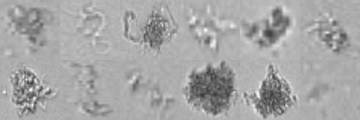}
\caption{Blue channel}
\end{subfigure}
\caption{An example set of subvisible protein aggregate images is shown in (a) RGB format, followed by (b) grayscale, and (c, d, e) individual red, green, and blue channels, respectively. Protein aggregates on the left side are formed due to heat stress, whereas those on the right are formed due to mechanical stress.}
\label{fig:SvP_images}
\end{figure}

\subsection{Materials}
\label{sec:materials}

We use eight monoclonal antibodies (mAbs) from commercial products that are commonly used to treat various cancer and immune diseases, and anonymize their product name from mAb1 to mAb8. For all experiments, we use chemicals that are pharmaceutical grade or higher.

\textbf{Heat stress}\,\textendash\,
We reformulate the commercial mAbs for heat stress using cation exchange chromatography and dialyze them first into 10 mM histidine buffer at pH 6 and then into 10 µM acetate buffer at pH 5. Following this, we adjust their concentration to 0.5 mg/ml using Amicon Ultra-15 Centrifugal Filters with a molecular weight cut-off of 30 kDa and filter the samples using a 0.22 µm PVDF syringe filter. Finally, we heat them in a digital heat block with and without the addition of 0.05\% polysorbate 20, in 2 ml microcentrifuge tubes to 90°C for 5 and 20 minutes

\textbf{Mechanical stress}\,\textendash\,
For mechanical stress, we dilute mAbs to 1 mg/ml directly from the commercial products using 0.9\% NaCl and 10 mM histidine buffer at pH 6 with 0.05\% polysorbate 20. We fill each protein (1.35 ml) in 2R glass vials under filtration with a 0.22 µm PVDF syringe filter. The samples are then shaken at 300 rpm for 48 hours at room temperature on a digital orbital shaker.

All aforementioned stress tests were performed in triplicates.

\textbf{Flow imaging microscopy}\,\textendash\,For imaging, we use FlowCam 8100 with VisualSpreadsheet software version 6.0.2.167 (Yokogawa Fluid Imaging Technologies, Scarborough, USA) and perform experiments in three replicates. Our setup consists of a 10$\times$ objective with an $80 \times 700$ \textmu m flow cell with the auto image frame rate of 27 frames/second and with the flow rate of 0.15 ml/min. We setup the device so that the particles are identified when pixel intensity exceeds 13 or 10 for light or dark pixels and when the distance between particles is $>$3 \textmu m. During the experiments the cleanliness of the flow cell is ensured by flushing it with water as well as detergent.

\subsection{Dataset}
\label{sec:dataset}

Among the images collected using the method described in Section~\ref{sec:materials}, we select 16,000 images larger than $25 \times 25$ pixels, with half subjected to heat stress and the other half to mechanical stress. The distribution of the resolutions for antibodies is shown in Figure~\ref{fig:scatter_plots}. This dataset was divided into training and validation sets, ensuring a balanced distribution of antibodies across both (refer to Table~\ref{tbl:dataset} for details). For consistency, each image was resized such that the largest edge measures 256 pixels, with the short edge padded using the median color of the image. Details on how these images are utilized during training are provided in Section~\ref{sec:training}. Compared to previous research efforts, our curated dataset stands out in two aspects.

\textbf{Shared antibodies with differing stress types}\,\textendash\,mAb1 and mAb2 antibodies include samples stressed with both methods, while the other antibodies are stressed with only one, allowing us to identify whether trained models can detect different stress types for the same antibody.

\textbf{Unseen antibodies in the training set}\,\textendash\,We deliberately exclude mAb5 and mAb8 from the training set to evaluate the generalization capability of the models on antibodies unseen during training. 

\subsection{Models}
\label{sec:models}

In this study, we employ two architectures that have seen extensive use in recent years: ResNet-50~\cite{resnet} and ViT-B/16~\cite{vit}. These models represent a convolutional neural network (CNN) and a vision transformer (ViT) model, respectively~\cite{lecun1998gradient,attention}. ResNet-50 is considered a lightweight model compared to ViT-B/16 and leverages deep residual learning to efficiently train very deep networks by introducing shortcut connections. On the other hand, ViT-B/16 utilizes the transformer-based architecture, which has been highly successful in natural language processing~\cite{gpt3}, and adapts it for image recognition tasks.

In recent years, the use of pretrained models has become a standard practice of modern machine learning and deep learning research in fields with limited data availability. These models, which have been trained on large datasets prior to being fine-tuned for specific tasks, offer numerous advantages that make them invaluable in a wide range of applications~\cite{ILSVRC15:rus}. Apart from supervised pretraining, recent research on self-supervised pretraining has shown remarkable potential in further enhancing the capabilities of these models~\cite{mocov1,simclr}. Self-supervised pretraining allows models to learn useful representations from unlabelled data by leveraging the inherent structure and relationships within the data itself. To take advantage of these developments, we employ the ResNet-50 model pretrained with supervised pretraining, Swav~\cite{swav}, MoCo-v3~\cite{moco_v3}, and VicReg~\cite{vicreg}. For ViT, we use supervised pretraining, DINO~\cite{dino}, and MAE~\cite{mae}. Detailed discussions on the benefits and shortcomings of these self-supervised frameworks can be found in their respective papers, as well as in the following surveys~\cite{khan2022contrastive_survey,ozbulak2023know}.

\subsection{Color Conversion}
\label{sec:color_conversion}

In order to observe discrepancies between color and monochrome FIM images, we obtain RGB images using the methodology outlined in Section~\ref{sec:materials} and convert those images into monochrome images. This approach allows us to maintain particle consistency in terms of morphology and distribution, enabling a direct comparison of the image features that may vary with color information.

To convert color SvP images (henceforth referred to as RGB images, representing red, green, and blue channels) into grayscale images, we employ the standard ITU-R 601-2 LUMA conversion method~\cite{luma}. This method calculates the luminance (L) as follows:
\begin{equation}
\text{L} = \text{R} \cdot \frac{299}{1000} + \text{G} \cdot \frac{587}{1000} + \text{B} \cdot \frac{114}{1000},
\end{equation}
where R, G, and B denote the values in their respective color channels.

In addition to the grayscale SvP images generated using this method, we also explore the use of individual RGB channels to further support the efficacy of colored FIM in our study. This approach allows us to analyze and compare the contributions of each color channel to the overall analysis of subvisible particles. An example set of SvP images in various color formats are provided in Figure~\ref{fig:SvP_images}. As can be seen, the differences across the images shown are subtle but noticeable on a case-by-case basis. 

\subsection{Training}
\label{sec:training}

We train all models using a grid-search approach, leveraging the SGD optimizer with four different learning rates -- 1, 0.1, 0.01, and 0.001 -- and a weight decay of either 0 or $10^{-5}$. We employ the commonly used Cross-Entropy Loss with a batch size of 32. Following research on downstream transferability for self-supervised models, we implement a cosine annealing learning rate scheduler for a total of 50 training epochs~\cite{simsiam}. We experiment with a momentum of 0, 0.1, and 0.05. Augmentations during training time include random resized crops of size $224 \times 224$, along with both vertical and horizontal flips, where the goal is to mitigate any spurious correlations that may arise due to padding. Each training run employs an early stopping strategy with a patience of 5 epochs.

\section{Experimental Results}
\label{sec:experimental_results}

The approach described in Section~\ref{sec:training} results in 24 training runs for each model and each color mode, comprising a large-scale experiment involving more than 800 training runs. In this section, we present a summary of the results obtained from these experiments and briefly discuss their implications.

\begin{table}[t]
\centering
\caption{The accuracy of the best-performing model for each color mode is provided, considering different pretraining methods. In each row, the highest accuracy is highlighted in \textbf{bold}, and the second highest is \underline{underlined}.}
\label{tbl:overall_accuracy}
\begin{tabular}{ccccccc}
\toprule
~ & ~ & \multicolumn{5}{c}{Color mode} \\
\cmidrule[0.75pt]{3-7}
\phantom{--}Model\phantom{--} & \phantom{--}Pretraining\phantom{--} & \phantom{--}RGB\phantom{--} & \phantom{--}\red{Red}\phantom{--} & \phantom{--}\green{Green}\phantom{--} & \phantom{--}\blue{Blue}\phantom{--} & \phantom{--}\gray{Grayscale}\phantom{--} \\
\midrule
\multirow{4}{*}{ResNet-50} & Supervised & \textbf{96.1}  & 94.0 & 95.1 & 93.6 & \underline{95.1} \\
~ & Swav & \textbf{97.0} & 94.0 & \underline{95.6} & 93.4 & 95.2 \\
~ & MoCo-v3 & \textbf{96.9} & 93.7 & 95.0 & 93.5 & \underline{96.2} \\
~ & VicReg & \textbf{97.0} & 94.2 & \underline{95.4} & 93.3 & 95.1 \\
\midrule
\multirow{4}{*}{ViT-B/16} & Supervised & \textbf{96.6} & 94.4 & \underline{95.8} & 94.3 & 95.7 \\
~ & DINO & \textbf{96.1} & 94.2 & \underline{95.4} & 93.2 & 95.1 \\
~ & MAE & \textbf{96.0} & 94.3 & \underline{95.0} & 93.6 & 94.9 \\
\bottomrule
\end{tabular}
\end{table}

\textbf{Overall performance}\,\textendash\,In Table~\ref{tbl:overall_accuracy}, we present the accuracy of the best-performing models for the five color modalities considered in this study, highlighting the highest accuracy for each row in bold. As can be seen, models trained with RGB images consistently outperform those trained using other color modalities, demonstrating the usefulness of color in detecting stress types that lead to the creation of SvPs. For the best performing models, namely the Swav- and VicReg-pretrained models, the difference between the RGB mode and the second-best mode is roughly $1.5\%$. On its own, this increase might seem insignificant, but having an overall accuracy increase of $1.5\%$ at $95.5\%$ means that approximately $33\%$ of the particles that would otherwise be incorrectly classified are now classified correctly.

\textbf{Self-supervised models}\,\textendash\,We find that self-supervised pretrained models are comparable to supervised pretrained models for both architectures. In particular, self-supervised learning models on ResNet-50 demonstrate superior performance compared to those with supervised pretraining. We suggest that further research efforts focusing on SvPs can leverage this advancement to enhance model performance.

\textbf{Antibody analysis}\,\textendash\,Using the best-performing model for each color mode, we investigate the true positive rates (TPR) for each antibody by calculating the proportion of antibody images correctly classified by the model. These results are presented in Table~\ref{tbl:antibody_tpr}. Interestingly, we find that the overall model performance is not uniform across different antibodies, indicating that it is easier to identify stress types for certain antibodies than for others. In particular, we observe that detecting stress types for mAb1, mAb2, mAb3, and mAb6 is easier compared to other antibodies. On the other hand, our models appear to accurately identify different stress types for the same antibody, as we observe high TPR rates for both mAb1 and mAb2. Furthermore, our models can also detect stress types for antibodies they are not trained with, as shown by the results for mAb5 and mAb8.

\textbf{Mixed-color training}\,\textendash\,Table~\ref{tbl:antibody_tpr} allows making another interesting observation: although the model that is trained using the RGB color mode has the overall best performance, when the TPRs are investigated for individual antibodies, several other color mode have a higher accuracy for certain antibodies. Prompted by this observation, we perform the same grid-search training routine with a mixed-color scheme where we use color mode as a method of data augmentation. That is, we randomly convert the color mode of an image from RGB into one of the individual colors. Doing so, we were able to train a final model that achieves an overall accuracy of $97.1\%$, with individual antibody TPRs listed in the bottom row of Table~\ref{tbl:antibody_tpr}. These results indicate that, unlike natural images where the color often does not play a crucial role, for SvP-related tasks, the usage of color as a an augmentation has the potential to make a noticeable difference in performance.

\begin{table}[t]
\centering
\caption{Using the overall best-performing model for each color mode, the TPR for each antibody is provided, categorized into (top) heat stress antibodies and (bottom) mechanical stress antibodies. The highest and the second highest TPR across models is highlighted using \textbf{bold} and \underline{underlined} text, respectively.}
\label{tbl:antibody_tpr}
\scriptsize
\begin{tabular}{ccccccc}
\toprule
\multirow{2}{*}{\shortstack{\\Color\\mode}} & \multirow{2}{*}{\shortstack{\\Model\\(pretraining)}} &  \multicolumn{5}{c}{Heat stress} \\
\cmidrule[0.75pt]{3-7}
~ & ~ & \phantom{-}mAb1\phantom{-} & \phantom{-}mAb2\phantom{-}  & \phantom{-}mAb3\phantom{-} & \phantom{-}mAb4\phantom{-} &  \phantom{-}mAb5\phantom{-}\\
\midrule
RGB & R-50 (Swav) & \textbf{97.3} & \underline{99.7} & \textbf{100} & \underline{94.4} & \textbf{97.0} \\
\red{Red} & ViT-B (Sup.) & 91.9 & 95.0 & 99.2 & 91.1 & 91.7 \\
\green{Green} & ViT-B (Sup.)  & 93.0 & 98.2 & 100 & 91.1 & 94.1 \\
\blue{Blue} & ViT-B (Sup.)  & 90.3 & 93.8 & 100 & 88.8 & 89.2 \\
\gray{Grayscale} & R-50 (MoCo) & 93.5 & 97.7 & 99.2 & 91.1 & 92.1 \\
\midrule
Mixed color & R-50 (Swav) & \underline{96.2} & \textbf{100} & \textbf{100} & \textbf{94.8} & \underline{95.5}\\
\bottomrule
\\
\toprule
\multirow{2}{*}{\shortstack{\\Color\\mode}} & \multirow{2}{*}{\shortstack{\\Model\\(pretraining)}} &  \multicolumn{5}{c}{Mechanical stress} \\
\cmidrule[0.75pt]{3-7}
~ & ~ & \phantom{-}mAb1\phantom{-} & \phantom{-}mAb2\phantom{-}  & \phantom{-}mAb6\phantom{-} & \phantom{-}mAb7\phantom{-} & \phantom{-}mAb8\phantom{-} \\
\midrule
RGB & R-50 (Swav) & \textbf{99.3} & 95.4 & 97.7 & 94.7 & 93.1 \\
\red{Red} & ViT-B (Sup.) & 96.9 & 95.9 & 96.1 & 94.0 & 92.0 \\
\green{Green} & ViT-B (Sup.) & 97.2 & \underline{96.4} & 98.0 & 95.1 & 93.1 \\
\blue{Blue} & ViT-B (Sup.) & 97.4 & 94.4 & \textbf{98.7} & \underline{95.8} & \underline{93.7} \\
\gray{Grayscale} & R-50 (MoCo) & 97.9 & 96.4 & \underline{98.0} & 94.7 & \textbf{94.7} \\
\midrule
Mixed color & R-50 (Swav) & \underline{98.6} & \textbf{96.9} & 97.8 & \textbf{96.5} & 93.1\\
\bottomrule
\end{tabular}
\end{table}

\section{Conclusions and Future Perspectives}

In this study, we demonstrated the usefulness of recent advancements in color FIM for improving the performance of state-of-the-art deep neural networks in detecting stress types that cause the creation of subvisible protein aggregates in biopharmaceuticals. With large-scale experiments involving eight commercial drugs, we showed that models trained with RGB images consistently outperform models trained with monochrome images. Lastly, we experimented with a mixed-color training strategy, which further improves the overall performance of the models compared to training images with RGB color-mode.

Intriguingly, our experiments reveal that the green channel is noticeably more useful for stress source detection compared to the red and blue channels. This suggests that the green wavelength contains specific information that is particularly advantageous for identifying stress sources. We leave further investigation into the underlying reasons for this phenomenon to future research efforts.

Apart from the stress source detection, we believe that a detailed investigation of protein aggregate patterns with color FIM holds promise. Additionally, the studies could be extended to include non-protein particles present in protein therapeutics, such as silicone oil, glass, and air bubbles.

\section{Acknowledgements}
We thank the hospital pharmacy from UZ Ghent for providing samples. MC is a doctoral fellow from the Research Foundation-Flanders (FWO-V) (grant number 1SH1S24N-7021). HLS would like to acknowledge the funding from Ghent University (grant numbers BOF/BAS/2022/051 and BOF/STA/202109/034).

MC SK YL YJS and HLS provided conceptual ideas and contextualization. MC provided the images. UO performed the machine learning experiments and wrote the initial draft. All authors corrected and approved the final version of the manuscript.

\bibliographystyle{splncs04}
\bibliography{main}

\begin{thebibliography}{10}
\providecommand{\url}[1]{\texttt{#1}}
\providecommand{\urlprefix}{URL }
\providecommand{\doi}[1]{https://doi.org/#1}

\bibitem{arakawa2000protection}
Arakawa, T., Kita, Y.: Protection of bovine serum albumin from aggregation by tween 80. Journal of Pharmaceutical Sciences  \textbf{89}(5),  646--651 (2000)

\bibitem{vicreg}
Bardes, A., Ponce, J., LeCun, Y.: {VICReg}: Variance-invariance-covariance regularization for self-supervised learning. arXiv preprint arXiv:2105.04906  (2021)

\bibitem{gpt3}
Brown, T., Mann, B., Ryder, N., Subbiah, M., Kaplan, J.D., Dhariwal, P., Neelakantan, A., Shyam, P., Sastry, G., Askell, A., et~al.: Language models are few-shot learners. Advances in Neural Information Processing Systems  \textbf{33},  1877--1901 (2020)

\bibitem{swav}
Caron, M., Misra, I., Mairal, J., Goyal, P., Bojanowski, P., Joulin, A.: Unsupervised learning of visual features by contrasting cluster assignments. Advances in Neural Information Processing Systems  \textbf{33},  9912--9924 (2020)

\bibitem{dino}
Caron, M., Touvron, H., Misra, I., J{\'e}gou, H., Mairal, J., Bojanowski, P., Joulin, A.: Emerging properties in self-supervised vision transformers. In: Proceedings of the IEEE/CVF International Conference on Computer Vision. pp. 9650--9660 (2021)

\bibitem{simclr}
Chen, T., Kornblith, S., Norouzi, M., Hinton, G.: A simple framework for contrastive learning of visual representations. In: International Conference on Machine Learning. pp. 1597--1607. PMLR (2020)

\bibitem{simsiam}
Chen, X., He, K.: Exploring simple siamese representation learning. In: Proceedings of the IEEE/CVF Conference on Computer Vision and Pattern Recognition. pp. 15750--15758 (2021)

\bibitem{moco_v3}
Chen, X., Xie, S., He, K.: An empirical study of training self-supervised vision transformers. In: Proceedings of the IEEE/CVF International Conference on Computer Vision. pp. 9640--9649 (2021)

\bibitem{vit}
Dosovitskiy, A., Beyer, L., Kolesnikov, A., Weissenborn, D., Zhai, X., Unterthiner, T., Dehghani, M., Minderer, M., Heigold, G., Gelly, S., et~al.: An image is worth 16x16 words: Transformers for image recognition at scale. arXiv preprint arXiv:2010.11929  (2020)

\bibitem{mae}
He, K., Chen, X., Xie, S., Li, Y., Doll{\'a}r, P., Girshick, R.: Masked autoencoders are scalable vision learners. In: Proceedings of the IEEE/CVF Conference on Computer Vision and Pattern Recognition. pp. 16000--16009 (2022)

\bibitem{mocov1}
He, K., Fan, H., Wu, Y., Xie, S., Girshick, R.: Momentum contrast for unsupervised visual representation learning. In: Proceedings of the IEEE/CVF Conference on Computer Vision and Pattern Recognition. pp. 9729--9738 (2020)

\bibitem{resnet}
He, K., Zhang, X., Ren, S., Sun, J.: {Deep residual learning for image recognition}. In: Proceedings of the IEEE/CVF Conference on Computer Vision and Pattern Recognition (2016)

\bibitem{luma}
Kanan, C., Cottrell, G.W.: Color-to-grayscale: does the method matter in image recognition? PloS one  \textbf{7}(1),  e29740 (2012)

\bibitem{khan2022contrastive_survey}
Khan, A., AlBarri, S., Manzoor, M.A.: Contrastive self-supervised learning: a survey on different architectures. In: 2022 2nd International Conference on Artificial Intelligence (ICAI). pp.~1--6. IEEE (2022)

\bibitem{kiese2008shaken}
Kiese, S., Papppenberger, A., Friess, W., Mahler, H.C.: Shaken, not stirred: mechanical stress testing of an igg1 antibody. Journal of pharmaceutical sciences  \textbf{97}(10),  4347--4366 (2008)

\bibitem{lai2022machine}
Lai, P.K., Gallegos, A., Mody, N., Sathish, H.A., Trout, B.L.: Machine learning prediction of antibody aggregation and viscosity for high concentration formulation development of protein therapeutics. In: MAbs. vol.~14, p. 2026208. Taylor \& Francis (2022)

\bibitem{lecun1998gradient}
LeCun, Y., Bottou, L., Bengio, Y., Haffner, P.: {Gradient-based learning applied to document recognition}. Proceedings of the IEEE  (1998)

\bibitem{matter2019variance}
Matter, A., Koulov, A., Singh, S., Mahler, H.C., Reinisch, H., Langer, C., Zucol, B., Mathaes, R.: Variance between different light obscuration and flow imaging microscopy instruments and the impact of instrument calibration. Journal of Pharmaceutical Sciences  \textbf{108}(7),  2397--2405 (2019)

\bibitem{moussa2016immunogenicity}
Moussa, E.M., Panchal, J.P., Moorthy, B.S., Blum, J.S., Joubert, M.K., Narhi, L.O., Topp, E.M.: Immunogenicity of therapeutic protein aggregates. Journal of pharmaceutical sciences  \textbf{105}(2),  417--430 (2016)

\bibitem{mullard2021fda}
Mullard, A.: Fda approves 100th monoclonal antibody product. Nature reviews. Drug discovery  \textbf{20}(7),  491--495 (2021)

\bibitem{narhi2015subvisible}
Narhi, L.O., Corvari, V., Ripple, D.C., Afonina, N., Cecchini, I., Defelippis, M.R., Garidel, P., Herre, A., Koulov, A.V., Lubiniecki, T., et~al.: Subvisible (2-100 $\mu$m) particle analysis during biotherapeutic drug product development: part 1, considerations and strategy. Journal of pharmaceutical sciences  \textbf{104}(6),  1899--1908 (2015)

\bibitem{ozbulak2023know}
Ozbulak, U., Lee, H.J., Boga, B., Anzaku, E.T., Park, H., Van~Messem, A., De~Neve, W., Vankerschaver, J.: Know your self-supervised learning: A survey on image-based generative and discriminative training. arXiv preprint arXiv:2305.13689  (2023)

\bibitem{fim_nn}
Poozesh, S., Cannav{\`o}, F., Manikwar, P.: Sensitivity and uncertainty analysis of micro-flow imaging for sub-visible particle measurements using artificial neural network. Pharmaceutical Research  \textbf{40}(3),  721--733 (2023)

\bibitem{ripple2012protein}
Ripple, D.C., Dimitrova, M.N.: Protein particles: What we know and what we do not know. Journal of pharmaceutical sciences  \textbf{101}(10),  3568--3579 (2012)

\bibitem{ILSVRC15:rus}
Russakovsky, O., Deng, J., Su, H., Krause, J., Satheesh, S., Ma, S., Huang, Z., Karpathy, A., Khosla, A., Bernstein, M., Berg, A.C., Fei-Fei, L.: {ImageNet large scale visual recognition challenge}. International Journal of Computer Vision  \textbf{115}(3),  211--252 (2015)

\bibitem{sharma2010micro}
Sharma, D.K., King, D., Oma, P., Merchant, C.: Micro-flow imaging: flow microscopy applied to sub-visible particulate analysis in protein formulations. The AAPS journal  \textbf{12}(3),  455--464 (2010)

\bibitem{thite2022machine}
Thite, N.G., Ghazvini, S., Wallace, N., Feldman, N., Calderon, C.P., Randolph, T.W.: Machine learning analysis provides insight into mechanisms of protein particle formation inside containers during mechanical agitation. Journal of pharmaceutical sciences  \textbf{111}(10),  2730--2744 (2022)

\bibitem{thorlaksen2023vitro}
Thorlaksen, C., Schultz, H.S., Gammelgaard, S.K., Jiskoot, W., Hatzakis, N.S., Nielsen, F.S., Solberg, H., Foder{\`a}, V., Bartholdy, C., Groenning, M.: In vitro and in vivo immunogenicity assessment of protein aggregate characteristics. International Journal of Pharmaceutics  \textbf{631},  122490 (2023)

\bibitem{attention}
Vaswani, A., Shazeer, N., Parmar, N., Uszkoreit, J., Jones, L., Gomez, A.N., Kaiser, {\L}., Polosukhin, I.: Attention is all you need. Advances in neural information processing systems  \textbf{30} (2017)

\bibitem{walsh2022biopharmaceutical}
Walsh, G., Walsh, E.: Biopharmaceutical benchmarks 2022. Nature Biotechnology  \textbf{40}(12),  1722--1760 (2022)

\bibitem{wang2005protein}
Wang, W.: Protein aggregation and its inhibition in biopharmaceutics. International journal of pharmaceutics  \textbf{289}(1-2),  1--30 (2005)

\bibitem{zolls2012particles}
Z{\"o}lls, S., Tantipolphan, R., Wiggenhorn, M., Winter, G., Jiskoot, W., Friess, W., Hawe, A.: Particles in therapeutic protein formulations, part 1: Overview of analytical methods. Journal of pharmaceutical sciences  \textbf{101}(3),  914--935 (2012)

\end{thebibliography}
\end{document}